\documentclass[10pt,twocolumn,letterpaper]{article}

\usepackage[pagenumbers]{wacv}              
\usepackage[accsupp]{axessibility} 
\usepackage{pbalance}
\usepackage{enumitem}
\usepackage{caption}

\makeatletter
\renewcommand\paragraph{\@startsection{paragraph}{4}{\z@}%
  {1ex \@plus .2ex \@minus .1ex}%
  {-0.6em}%
  {\normalfont\normalsize\bfseries}}
\makeatother

\AtBeginDocument{%
  \setlength{\abovedisplayskip}{3pt plus 1pt minus 1pt}
  \setlength{\belowdisplayskip}{1pt plus 1pt minus 1pt}
  \setlength{\abovedisplayshortskip}{2pt plus 1pt minus 1pt}
  \setlength{\belowdisplayshortskip}{2pt plus 1pt minus 1pt}
}
\makeatletter
\renewcommand\subsection{\@startsection{subsection}{2}{\z@}%
  {-1.5ex \@plus -.2ex \@minus -.2ex}
  {0.4ex \@plus .1ex}
  {\normalfont\large\bfseries}}
\makeatother

\makeatletter
\def\wacvsection{\@startsection{section}{1}{\z@}%
  {-4pt \@plus -2pt \@minus -2pt}{3pt}{\large\bf}}
\def\wacvsubsection{\@startsection{subsection}{2}{\z@}%
  {-1pt \@plus -2pt \@minus -2pt}{2pt}{\normalfont\large\bfseries}}
\makeatother

\definecolor{wacvblue}{rgb}{0.21,0.49,0.74}
\usepackage[pagebackref,breaklinks,colorlinks,allcolors=wacvblue]{hyperref}

\title{CoDecide: Content-Adaptive Video Decoding for High-Throughput Video Analytics}

\author{Shristy Shah\\
Aalborg University, Denmark\\
Milestone Systems, Denmark\\
{\tt\small ssha@create.aau.dk}
\and
Thomas B. Moeslund\\
Aalborg University, Denmark\\
{\tt\small tbm@create.aau.dk}
\and
Kamal Nasrollahi\\
Aalborg University, Denmark\\
Milestone Systems, Denmark\\
{\tt\small kn@create.aau.dk}
}

\newcommand{\method}{CoDecide\xspace}
\AddToShipoutPicture*{\AtPageUpperLeft{\put(54,-30){\parbox[t]{\dimexpr\paperwidth-1.5in}{\centering\scriptsize \copyright~2027 IEEE. Personal use of this material is permitted. Permission from IEEE must be obtained for all other uses, in any current or future media, including reprinting/republishing this material for advertising or promotional purposes, creating new collective works, for resale or redistribution to servers or lists, or reuse of any copyrighted component of this work in other works.}}}}

\begin{document}
\maketitle
\begin{abstract}
Analytics on video archives are bottlenecked by decoding. In a dense pipeline that runs a detector on every frame, decoding is a small fraction of the runtime, but once the detector is restricted to selected frames, decoding takes over as the main cost. We introduce \method, a retrospective analytics system that only decodes the frames where new content appears, using signals from the compressed bitstream to find them. Motion tells us something is moving, but not whether it is new to the scene or has been there the whole time. To make that distinction, we supervise a lightweight compressed-domain student on a label-free novelty signal from a frozen DINOv2 teacher, which fires on semantically new regions rather than on any motion. Consequently, the student model generalizes to new camera deployments without requiring per-camera retraining. At inference, it decides which frames to decode and which regions to reconstruct. Detections from the decoded frames are propagated to the skipped frames using motion vectors. We evaluate \method with a leave-one-dataset-out protocol on VIRAT, MEVA, and ShanghaiTech. Against a dense per-frame detector, \method recovers more moving-object detections than prior compressed-domain selectors at an equivalent decoding budget, outperforming even a competitor trained directly on the target camera. \method runs $15$ to $24\times$ faster than a pipeline that decodes and detects every frame. Code is available at \href{https://github.com/shristy51/CoDecide}{github.com/shristy51/CoDecide}.\end{abstract}

\section{Introduction}
\label{sec:intro}
Retrospective video analysis runs vision models on large amounts of video recorded across many cameras over long durations. Common archival applications are modeling long-term traffic patterns in smart cities, analyzing customer behavior in retail environments, and conducting forensic reconstruction for security systems. These archives are analyzed repeatedly, as different downstream tasks evaluate the same footage over time. Decoding them once and caching is not an option, since that would require huge storage, and that is the reason the archive is compressed.

\begin{figure}[!t]
  \centering
  \includegraphics[width=\columnwidth]{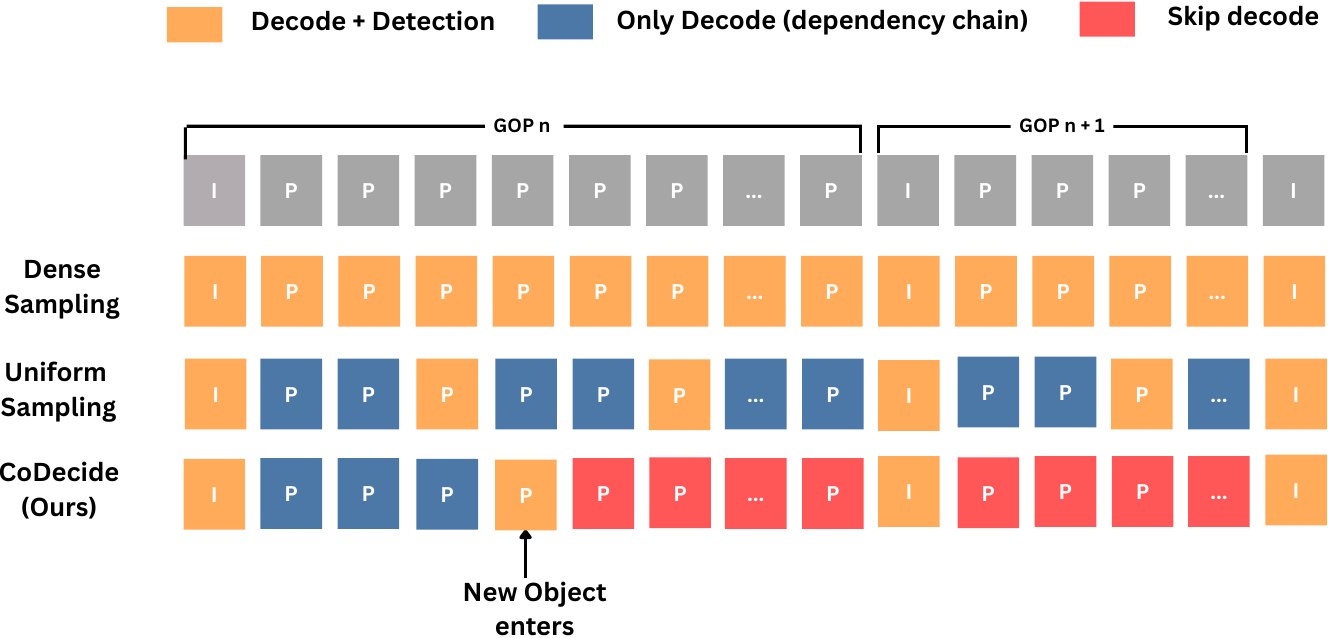}
  \caption{\textbf{Decoding strategies over a video stream.} Dense processing analyzes every frame. Uniform sampling reduces model inference but still traverses the codec dependency chain and risks missing sudden object arrivals. CoDecide (Ours) predicts novelty from compressed signals, decoding only up to the frame where new content appears.}
  \label{fig:teaser}
\end{figure}
This decoding of every frame to the pixel domain introduces a bottleneck, particularly in retrospective settings. Standard video codecs organize a video into chunks called groups of pictures (GOPs)~\cite{h264}, each beginning with an independently decodable intra-coded frame (I-frame) followed by dependent inter-coded P-frames, which reference past frames and optionally, bi-directionally predicted B-frames. I-frames have a large data footprint and contribute minimal novel information in highly static surveillance videos. Consequently, large-scale archival networks use longer GOPs to optimize storage efficiency. In our measurement on a 1080p H.264 video, the cost of reaching an inter frame accumulates with its depth in the GOP. For example, accessing the last inter-frame of a 250-frame GOP incurs roughly $45\times$ latency overhead compared to accessing the initial inter-frame.

Uniform sampling fails to save decoding cost because of this accumulation as shown in Figure~\ref{fig:teaser}. Within a GOP of length $L$, sampling at any stride shorter than $L$ still places a sample near the end of the chain, and reaching that sample decodes almost all $L$ frames.

To avoid running expensive vision models on every frame, prior works with database-driven video analytics engines use cascades of cheap specialized proxies or statistical filters to skip frames~\cite{noscope,balzeit,focus,probabilisticpredicates,tahoma,viva,eva}. Frame selection methods select a subset of frames that is important for a given downstream task and skip the rest~\cite{adaframe,korbar2019scsampler,zhi2021mgsampler,leap}. With few exceptions, both approaches still operate on fully decoded RGB frames.

To compress a frame, the encoder records how each block of pixels moved since the last frame, and flags the blocks it could not predict from what came before. Yet, pure motion is not equivalent to new semantic content. An object that entered the camera feed may continue moving for hundreds of frames, producing continuous motion vectors and residuals even though its visual identity is no longer new. The key challenge is determining whether a structural change represents content that has not been observed.

The proposed model, CoDecide, treats selective decoding as novelty prediction in the compressed domain. In archival footage, most GOPs contain no new content, so traversal can stop at the I-frame; the cost is worth paying only where something arrives.~For the tasks that archives are queried with, such as object detection, counting, and binary predicates, pixels are needed only when something new arrives; in between, motion vectors already describe how known objects move and can be used to propagate the results. We define novelty offline using patch features from a frozen DINOv2 backbone~\cite{dinov2}, marking a region novel when its representation finds no match in a semantic memory of what has already been decoded.~This definition is used for supervision of the lightweight compressed-domain~model.

We implement CoDecide on a modified FFmpeg~\cite{ffmpeg} decoder that exposes per-macroblock motion vectors, non-zero coefficient counts, and macroblock types during entropy decoding, and that supports reconstructing a selected subset of macroblocks. Full reconstruction is performed in two cases: for I-frames, which have no dependencies, and for frames that the predictor flags as novel. We evaluate our system in terms of accuracy and throughput while performing three archival tasks.

\noindent Our contributions are as follows:
\begin{itemize}[leftmargin=*,topsep=2pt,itemsep=1pt,parsep=0pt]
\item \textbf{Label-free semantic novelty.} We define novelty within a GOP as dissimilarity in DINOv2 feature space against what has already been decoded. Because supervision uses no class labels, detector outputs, or per-camera background models, one trained student transfers across surveillance datasets.
\item \textbf{Codec-domain student with memory.} We show that motion vectors, non-zero coefficient counts, and macroblock types are jointly useful to separate arrivals from persistent motion when read through a recurrent predictor that carries GOP-level history.
\item \textbf{Spatio-temporal selective decoding.} We introduce a policy that stops chain decoding at the last frame flagged novel and reconstructs only the macroblocks selected by the presence head, with detections propagated along motion vectors in between.
\end{itemize}

\section{Related Work}

\label{sec:related}

\paragraph{Efficient video analytics and frame selection.}
Prior work uses lightweight filters to select frames for heavy downstream models; however, nearly all rely on decoding fully reconstructed frames. NoScope~\cite{noscope} trains a smaller specialized model to imitate the target architecture and drops static frames via a difference detector. To handle complex queries, probabilistic predicates~\cite{probabilisticpredicates} learn optimized proxy filters that guarantee user-defined accuracy targets without requiring per-query retraining. Meanwhile, BlazeIt~\cite{balzeit} uses neural networks as control variates to accelerate approximate object counting. Tahoma~\cite{tahoma} works by optimizing the input representation (e.g., resolution and color depth) to maintain accuracy using the cheapest viable representation. Focus~\cite{focus} proposes indexing the objects in a video by running a network once over the whole archive and using this information for later queries. Frame selection in the video recognition literature chooses which subset of frames reaches the classifier. AdaFrame~\cite{adaframe} uses a recurrent memory network to sample frames adaptively for fast video recognition. MGSampler~\cite{zhi2021mgsampler} samples by motion rather than by time, giving busier parts of a video more frames. OCSampler~\cite{ocsampler} skims every candidate frame at a lower resolution before picking a subset in one shot for the full-resolution classifier. SCSampler~\cite{korbar2019scsampler} selects a few salient clips that are sufficient to recognize the action in a whole video. Because a variant of this method operates directly on compressed data, rejected clips are never decoded. However, the underlying task semantics differ fundamentally. Video recognition requires a single label for an entire video sequence, whereas retrospective analysis must account for events across every frame.

\begin{figure*}[!t]
\centering
\includegraphics[width=\textwidth]{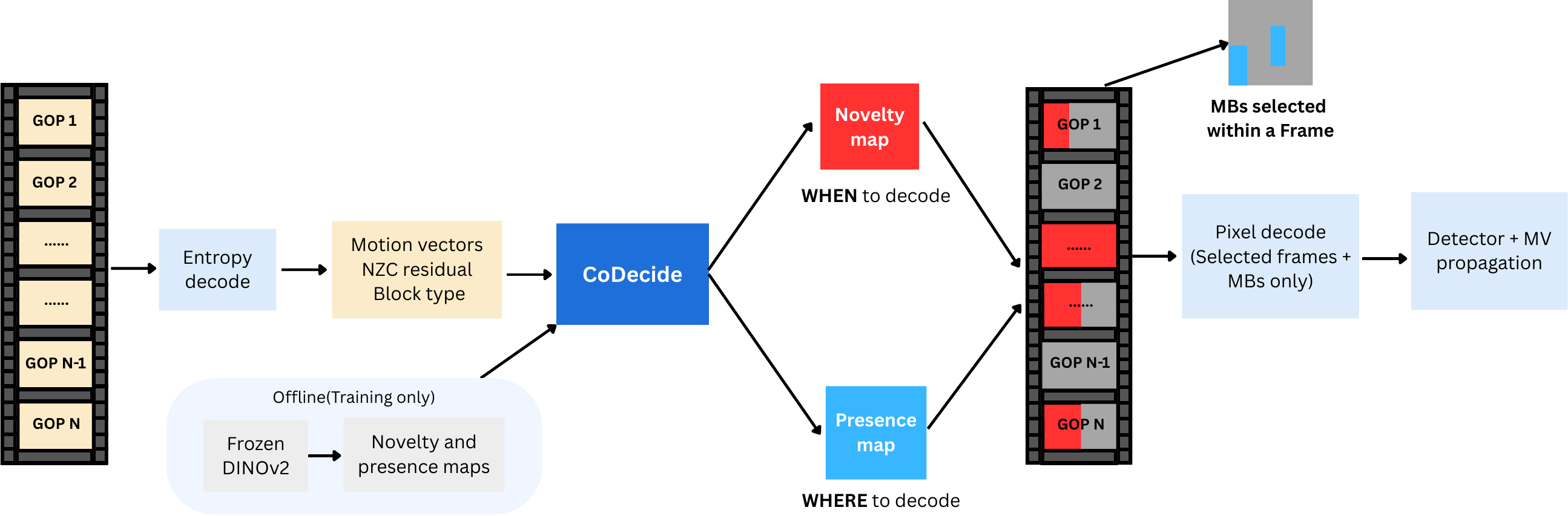}
\caption{\textbf{\method{} system architecture.} An incoming video bitstream is entropy-decoded into compressed features (MVs, NZC, macroblock types). The CoDecide network uses these to generate presence and novelty maps, filtering the video both spatially and temporally (indicated by grayed-out regions). Only selected un-grayed frames undergo pixel decoding and object detection, with bounding boxes propagated to skipped frames using motion vectors.}
\label{fig:pipeline}
\end{figure*}

\paragraph{Compressed-domain video analysis.}
Prior works have explored using codec signals as model input rather than as a filter. CoViAR~\cite{coviar} proposes to add motion vectors and residuals as input along with I-frames for action recognition. It uses accumulated motion vectors back to the I-frame so each frame can be handled on its own. DMC-Net~\cite{dmcnet} works on predicting optical flow using motion vectors and residuals as optical flow is computationally expensive.

CompViT~\cite{compvit} uses an asymmetric transformer network with I-frames for appearance features and motion vectors and residuals for motion features. In surveillance, motion vectors have long been used for moving-object detection and tracking~\cite{poppe2009moving,kas2009approach,szczerba2009fast,yokoyama2009motion,jaballah2019fast}, usually by thresholding or clustering the vector field on the macroblock grid. DMC-Net is the closest of these to our design as it is a cheap codec-domain network trained against a signal too expensive to run at deployment. In all of this work, the codec signal feeds the task model directly, whereas we use it to decide which frames the task model should see in full.

\paragraph{Decode-aware video systems.}
CoVA~\cite{cova} is closest to our objective. It learns background modeling in
the compressed domain by training a small U-Net on labels from a
Mixture-of-Gaussians background model. It then groups the predicted regions into blobs and tracks them using SORT. Every new track triggers one anchor frame to be decoded, propagating the detection along the track. This network is trained at query time for each video and does not transfer to a camera it was not trained on. PacketGame~\cite{packetgame} uses bitstream metadata (packet sizes and picture types) and learns to predict which packets a downstream model needs to maintain accuracy. Its predictor is trained from inference records for one task, and degrades when it is applied to a domain it was not trained on. Furthermore, packet size provides no spatial information regarding where content resides; consequently, the filtering decision is strictly temporal. ParaDeco~\cite{paradeco} examines systems like CoVA and observes that selecting few frames saves little decoding when those frames occupy late positions within a GOP, since all preceding frames must still be reconstructed. To eliminate this dependency, ParaDeco roots motion vectors back to an intra-coded block, measures frame contribution using these rooted vectors, and replaces selected inter-frames with pseudo-frames generated by a conditional GAN~\cite{pix2pix}.

\paragraph{Self-supervised features for label-free supervision.}

Self-supervised vision models learn features that carry semantic structure without class labels. DINO~\cite{dinopaper} showed that the features of a self-supervised ViT contain explicit information about the semantic segmentation of an image, and Amir et al.~\cite{amir2021deep} showed that DINO's patch descriptors can be used directly, without training, for co-segmentation, part co-segmentation, and semantic correspondence. DINOv2~\cite{dinov2} extends this to features that transfer across image distributions without finetuning, at the pixel level as well as the image level. We use this feature space to define novelty offline without deciding in advance what kind of objects we are looking for.
\section{Methodology}
\label{sec:method}
\paragraph{Background on codecs.}
As described in Section~\ref{sec:intro}, video encoding works by grouping consecutive frames into Groups of Pictures (GOPs) to reduce redundancy. In the H.264/AVC~\cite{h264} standard, each frame is divided into a grid of $16{\times}16$ pixel blocks called macroblocks. Instead of full pixel values, these macroblocks store a variety of low-level syntax elements that are available immediately after entropy decoding. For our pipeline, the key features of interest are motion vectors, which specify displacement relative to the closest match in a reference frame; the residual, which captures the remaining prediction error; and the macroblock type, which specifies how the block was encoded. Inter blocks are predicted using the reference frame, while intra blocks are directly coded, and static blocks containing no unique data are flagged as skip. Extracting these features requires no pixel-domain information. Newer codecs like HEVC~\cite{hvec} and VVC~\cite{vvc} provide similar information, but they organize it using tree-structured blocks rather than fixed macroblocks. We focus on H.264/AVC here because it is still the most widely deployed video codec.

\begin{figure*}[!t]
  \centering
  \includegraphics[width=\textwidth]{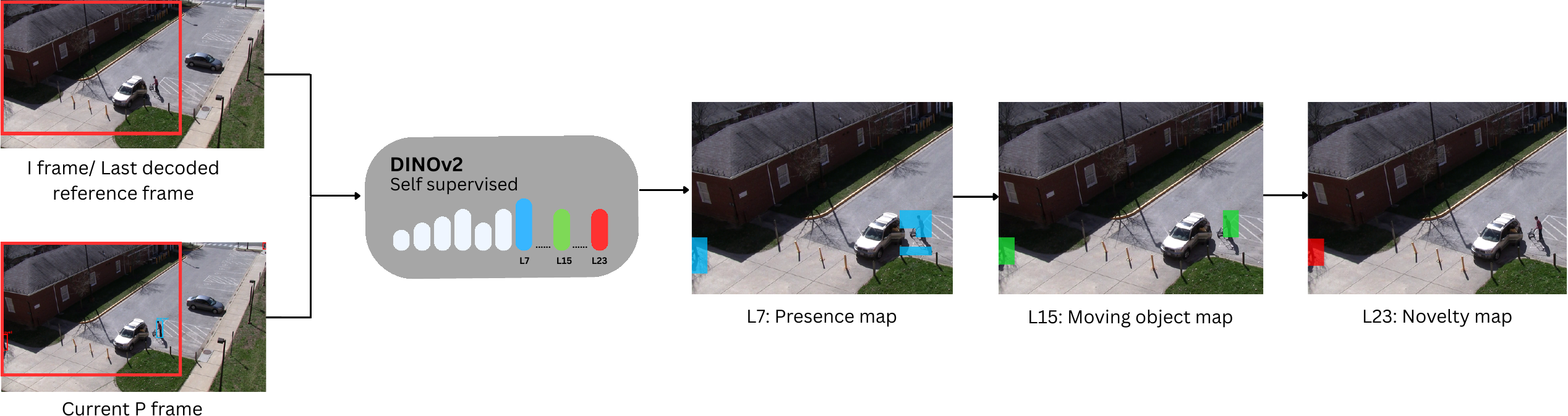}
  \caption{\textbf{The teacher's three stages.} Layer~7 dissimilarity against the
reference set marks everything that changed (Presence map). Layer~15 compares
those patches against the static background of the same frame and drops the
shadow, which is the same material as the road. Layer~23 matches what remains
against the reference one-to-one, and the patches left without a counterpart are
novel (Novelty map).}
  \label{fig:teacher-map}
\end{figure*}
\subsection{System Overview}
Figure~\ref{fig:pipeline} illustrates the system architecture of \method, which consists of three main components: a teacher model, a student network, and a decode policy. During offline training, the teacher measures frame-level novelty using a frozen DINOv2 backbone without manual annotation. At runtime, only the student network and decode policy are active. For each GOP, the bitstream is entropy-decoded to extract a macroblock-level feature grid containing motion vectors, non-zero coefficient counts (NZC), and macroblock types without performing any pixel-level reconstruction.

From these compressed-domain features, the student network predicts two outputs: a novelty map to identify frames containing new content, and a presence map to mark macroblocks whose content has changed since the I-frame. Both maps are required because the downstream detector must evaluate all active content within a reconstructed frame, not just newly arrived objects. The decode policy then evaluates the student's predicted frame-level novelty score against a threshold $\theta$. This threshold is user-tunable; adjusting $\theta$ sets the frame filtration rate and thus the trade-off between accuracy and compute. 

When the predicted novelty score exceeds $\theta$, the frame is marked as novel, and decoding is triggered up to the farthest novel frame in the GOP. Within this decoded chain, \method performs selective reconstruction: we decode only the macroblocks flagged by the presence map and any intra-coded macroblocks that lack a temporal reference, while populating the remaining regions via co-located blocks from the reference frame. Finally, the object detector runs on the I-frame and the selected novel frames. The resulting bounding boxes are then propagated to the skipped frames using their underlying motion vectors. If no frame in a GOP exceeds $\theta$, only the I-frame is decoded, and its detections are propagated across the entire GOP.

\subsection{Setup}
\label{sec:method:setup}
We re-encode all video sequences with a GOP length of $L{=}250$ frames, matching the standard archival footprint~\cite{cova}. The I-frame of each GOP is always decoded. For each P-frame, the extracted macroblock-level feature grid contains four discrete numerical channels: two representing the orthogonal motion vector components ($MV_x$ and $MV_y$), one tracking the non-zero coefficient counts (NZC), and one specifying the macroblock type. The initial spatial dimensions of this four-channel grid scale directly with the source resolution, varying across our datasets.

In contrast, the teacher network operates on a fixed spatial topology. Its novelty targets are derived from the patch tokens of a frozen DINOv2~\cite{dinov2} ViT-L/14 backbone fed with \(336\times336\) resolution images, which align to a \(24\times24\) grid. To optimize the student network against these spatial targets during training, we average-pool the four macroblock channels down to this matching $24\times24$ grid and temporally downsample the GOP into $T{=}32$ bins of approximately eight frames each. This spatial pooling standardizes the input dimensions, enabling a single student architecture to evaluate all datasets regardless of native resolution. At inference time, the student's predicted maps are upscaled back to the native macroblock resolution to guide the selective reconstruction pipeline.

\subsection{Label-free novelty teacher}
\label{sec:method:teacher}
The aim of the label-free novelty teacher is to isolate semantic novelty from transient pixel variations such as illumination shifts, shadows, sensor noise, or simple object displacement. We use a frozen DINOv2~\cite{dinov2} ViT-L/14 backbone as our feature extractor. Unlike architectures like CLIP~\cite{radford2021learning} which is optimized for caption alignment or MAE~\cite{he2022masked} whose features are trained for pixel reconstruction, DINOv2's self-supervised training yields dense features with strong empirical performance on downstream semantic benchmarks~\cite{dinov2}. Consequently, the cosine similarity between its patch tokens aligns with dense semantic correspondences rather than surface textures, as earlier shown for DINO features~\cite{amir2021deep}.

\Cref{fig:teacher-map} illustrates the three steps used by the teacher to measure novelty for a given frame within a GOP, utilizing features extracted from layers 7, 15, and 23. Given a normalized patch token $f^\ell_i(t)$ from layer $\ell$ of frame $t$ and a step-specific set of normalized reference tokens $\{r^\ell_j\}$, the local patch dissimilarity is scored as:
\begin{equation}
  d^\ell_i(t) \;=\; 1 - \max_j\, \big\langle f^\ell_i(t),\, r^\ell_j \big\rangle .
  \label{eq:dissimilarity}
\end{equation}
The maximization step identifies the most semantically corresponding token from the reference set using cosine similarity, where $j$ indexes through all tokens in the reference set regardless of spatial location.

The shallow layer, Layer 7, describes low-level visual appearances. When this layer is compared against the I-frame of the GOP, all dissimilar patches are flagged. This defines the presence map region $R_7$.  Beyond the foreground object, $R_7$ captures contextual background alterations like moving shadows or newly uncovered regions.

Layer 15 is subsequently used to filter out these transient background elements, isolating the core moving objects into a refined object region, $R_{15}$. 
At this depth, features represent abstract object concepts. Using the static regions outside $R_7$ as the reference set, newly uncovered or shadowed background patches find semantic matches and are discarded, while unmatched moving foreground objects are retained. This refined set of dynamic object patches constitutes $R_{15}$.

Finally, these isolated object patches are grouped into spatially connected regions, where Layer 23 features are used to evaluate true semantic novelty, yielding the final novelty region $R_{23} \subseteq R_{15}$. 
At this deep layer, features capture semantic identity, and the reference set consists of the features of the same region in the I-frame and prior novel frames. To isolate true structural novelty, the Hungarian algorithm~\cite{kuhn1955hungarian} enforces a strict one-to-one patch mapping against historical references, preventing new objects from duplicating matches.

At this final stage, the token-level dissimilarity values are aggregated into a region-level novelty score, defined as the fraction of patches within the connected region that lack a high-confidence semantic match. 
Layer 15 processing is essential to this final alignment; without it, newly uncovered background patches would enter Layer 23, compete with the banked object patches under the Hungarian assignment, and record mismatches. Consequently, an object leaving the frame would incorrectly trigger a novelty alert.

\paragraph{Bank-on-decode.}
When a region triggers a novelty event in a GOP, that frame is appended to the reference set. As a result, subsequent frames successfully match against the banked object, suppressing redundant alerts. This design ensures the teacher emits a single pulse near an object's arrival, satisfying the requirements of the downstream decode policy. The reference set resets to the I-frame at the beginning of each GOP. All presence and novelty maps are pre-computed in a single offline pass and cached. All hyperparameter thresholds are shared across all datasets and are detailed in the supplementary material.

\begin{figure*}[!t]
\centering
\includegraphics[width=\textwidth]{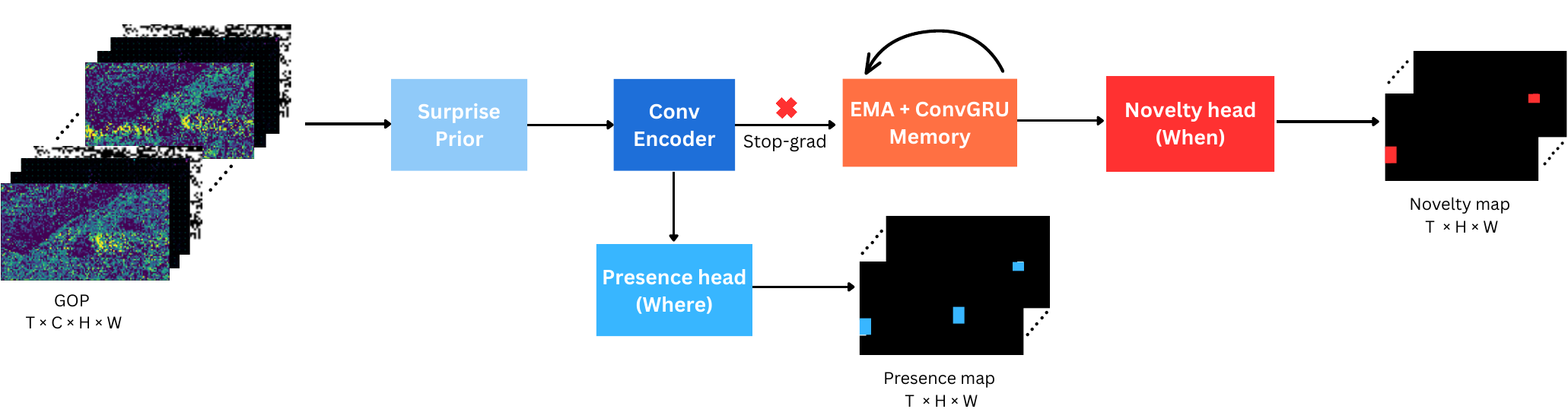}
\caption{\textbf{Student architecture.} The encoder maps each temporal bin to features on its own. The presence head predicts where content changed. The novelty head compares the current bin against an EMA baseline and a ConvGRU memory, both started from the I-frame bin, so it responds to arrival rather than to presence.}
\label{fig:student}
\end{figure*}
\subsection{Compressed-domain student}
\label{sec:method:student}

\Cref{fig:student} shows the student model. It takes the four macroblock channels as input and learns to predict the same two maps as the teacher. These labels are sparse because a GOP contains hundreds of frames, and the teacher fires once, or not at all, per GOP.

We derive a fifth input channel, the surprise prior, from these four. Motion on its own says little about novelty, because the codec produces motion vectors for anything that shifts, including content it went on to predict well. What is really important is the motion that cannot be explained by the codec. This is achieved by multiplying the magnitude of movement by the residual proxy. This isolates blocks that move but are poorly predicted. The presence head predicts where content changed in each bin. The teacher's presence map is the target for this head.

Novelty cannot be predicted using just the current bin. The novelty head calculates the drift of the bin from the two references: a non-learning exponential moving average (EMA)~\cite{ema} and a recurrent ConvGRU~\cite{convgru}, which decides what has to be kept in memory. They both have the same initial state as the I-frame bin; hence, the drift at the beginning of the GOP is zero. The drift rises when a new object enters and falls back as the references absorb it. 
Gradients from the novelty head do not reach the shared encoder. Letting the sparse novelty signal shape the features destabilized them, so the encoder is trained by the presence loss alone. The EMA is detached as well, so drift from it falls only when the features themselves stop changing. The ConvGRU, in contrast, learns from the novelty loss how quickly to absorb a new object. Every bin depends only on the bins before it, so running a whole GOP in one pass
gives the same result as feeding the bins one at a time.

\subsection{Distillation objective}
\label{sec:method:loss}

The distillation objective minimizes
\begin{equation}
\mathcal{L} \;=\; \mathcal{L}_{\text{pres}} + \mathcal{L}_{\text{nov}} + \lambda_s \mathcal{L}_{\text{sparse}} + \lambda_g \mathcal{L}_{\text{gop}} .
\label{eq:total}
\end{equation}

\noindent \(\mathcal{L}_{\text{pres}}\) and \(\mathcal{L}_{\text{nov}}\) are cross-entropy losses over both maps. Within \(\mathcal{L}_{\text{nov}}\), we heavily weight bins containing content but lacking novelty to penalize the false alerts the teacher avoids. $\mathcal{L}_{\text{sparse}}$ is an $L_1$ penalty on the novelty output, which stops the head staying on once it has fired.

$\mathcal{L}_{\text{gop}}$ compares a smooth maximum of the novelty map over the GOP against whether there was an arrival in this GOP at all, so the decode decision is supervised directly. The weights are in the supplementary.

\section{Experiments}
\label{sec:experiment}
\paragraph{Datasets.}
We evaluate on three surveillance datasets: VIRAT~\cite{virat} (parking lots and roads), ShanghaiTech~\cite{shanghaitech} (campus pedestrian pathways), and MEVA~\cite{meva} (multi-view vehicular and pedestrian traffic). These scenes consist of stationary vehicles, fast, non-rigid human motion, and standard traffic. ShanghaiTech is evaluated as a standard surveillance dataset rather than an anomaly benchmark, as its busier scenes heavily stress the system compared to VIRAT or MEVA.

\paragraph{Evaluation protocol.}
We use a strict leave-one-dataset-out (LODO) protocol. We trained three students with the same configuration (architecture and augmentations), with one dataset held out. This is done to check whether the unsupervised novelty signal generalizes across scenes and object categories without per-view training. Because novelty distributions vary by dataset, a static threshold yields inconsistent filtration rates. To compare under identical resource constraints, we calibrate $\theta$ on each held-out dataset to match the decode budget of the competitors. This uses only the student's scores, with no labels or accuracy, so a new camera can be calibrated the same way without retraining.

\paragraph{Reference pipeline.}
The upper limit is dense detection, YOLO11x~\cite{yolo11,ultralytics} on each frame. We compute fidelity, defined as the percentage of dense-detector boxes recovered by the selective pipeline, on all datasets.

\paragraph{Tasks and metrics.}
We measure the performance on three downstream tasks. For object detection, the metric is moving-object recall using IoU of $0.5$; for the binary predicate task, it is the per-frame accuracy of predicting whether a moving object exists, and for the counting task, it is the accuracy of counting the moving objects. We report only moving-object metrics because static objects are detected identically by all methods at the I-frame. Cost is measured using \textbf{Sel}, the fraction of frames sent to the detector, and \textbf{Dec}, the fraction of the GOP the codec must decode.

\paragraph{Implementation.}

All experiments are conducted using a single machine equipped with an NVIDIA RTX 2000 Ada Generation GPU, an Intel Core Ultra 9 285 CPU, and 125 GB RAM. The student model is trained using AdamW for 40 epochs, while the cache of the teacher is pre-computed before training. Entropy decoding during inference is done by our modified FFmpeg decoder.

\begin{table*}[!t]
\centering
\setlength{\tabcolsep}{4pt}
\renewcommand{\arraystretch}{1.15}
\resizebox{\textwidth}{!}{%
\begin{tabular}{@{}l ccccc ccccc ccccc@{}}
\toprule
& \multicolumn{5}{c}{VIRAT} & \multicolumn{5}{c}{ShanghaiTech} & \multicolumn{5}{c}{MEVA} \\
\cmidrule(lr){2-6}\cmidrule(lr){7-11}\cmidrule(l){12-16}
Method & OD & BP & Cnt & Sel & Dec & OD & BP & Cnt & Sel & Dec & OD & BP & Cnt & Sel & Dec \\
\midrule
Keyframes only $+$ prop. & 0.65 & 0.99 & 0.94 & 0.4\% & 0.4\% & 0.43 & 0.92 & 0.62 & 0.4\% & 0.4\% & 0.73 & 0.99 & 0.88 & 0.4\% & 0.4\% \\
CoVA~\cite{cova} $+$ prop. & 0.75 & 1.00 & 0.95 & 0.7\% & 30.0\% & 0.54 & 0.95 & 0.68 & 1.4\% & 45.7\% & 0.76 & 0.99 & 0.89 & 0.7\% & 17.3\% \\
PacketGame~\cite{packetgame} $+$ prop. & 0.77 & 1.00 & 0.96 & 3.5\% & 30.0\% & 0.64$^\dagger$ & 0.96 & 0.74 & 16.8\% & 40.0\% & 0.76 & 0.99 & 0.90 & 1.0\% & 15.2\% \\
ParaDeco$^*$~\cite{paradeco} $+$ prop. & 0.75 & 1.00 & 0.95 & 0.8\% & 51.4\% & 0.43 & 0.92 & 0.62 & 0.8\% & 36.9\% & 0.81$^\dagger$ & 0.99 & 0.92 & 0.8\% & 51.8\% \\
\midrule
\textbf{\method (ours)} & \textbf{0.78} & 1.00 & 0.95 & 1.0\% & 29.4\% & 0.60 & \textbf{0.98} & 0.74 & 1.9\% & 45.5\% & \textbf{0.78} & 0.99 & 0.92 & 0.7\% & 15.5\% \\
\bottomrule
\end{tabular}}
\caption{\textbf{Fidelity to dense detection} across datasets and tasks, read at
comparable decode budget. All numbers are the fraction of dense YOLO11x output
the selective pipeline reproduces, so $1.0$ is the ceiling everywhere.
``$+$\,prop.'' pairs each selector with our box propagation. $\dagger$ marks an entry above \method, but with much higher cost.
$^*$Our reimplementation without Frame Predictor.}
\label{tab:main}
\end{table*}

\begin{figure*}[t]
\centering
\includegraphics[width=\textwidth]{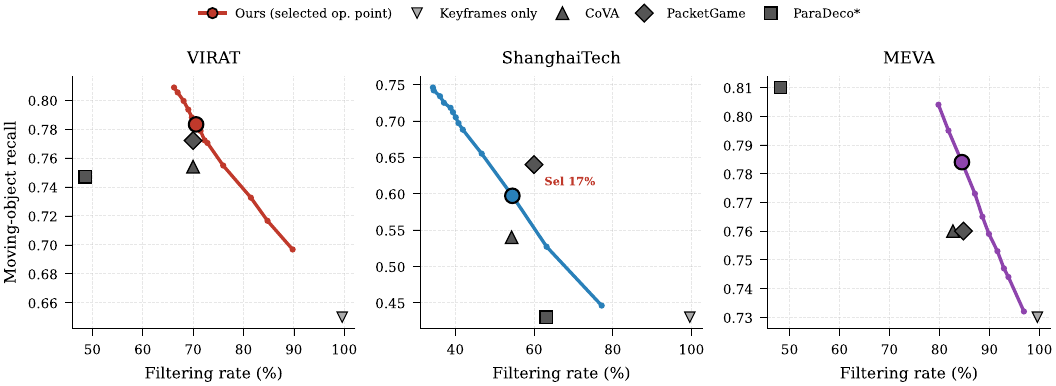}
\caption{\textbf{Recall vs.\ Filtration rate.} Each point on our curve
(colored) corresponds to a threshold $\theta$; the bold marker is the
operating point from \Cref{tab:main}. Baselines (gray) are plotted at their
reported chain-decode cost.}
\label{fig:filtering}
\end{figure*}

\subsection{Comparison with baselines}
\label{sec:exp:baselines}
All methods use the same detector (YOLO11x), the same cost accounting, and our motion vector propagation (`$+$\,prop.'), and so \Cref{tab:main} compares selection policies rather than the original systems.

`\emph{Keyframes only}' decodes the I-frame and propagates using the detections from that frame alone, which is the baseline with no selection. `\emph{CoVA}'~\cite{cova} uses tracking information of moving blobs in the compressed domain based on a per-camera background model. `\emph{PacketGame}'~\cite{packetgame} applies a trained predictor to metadata of the bitstream, and gates packets before decoding based on predicted importance. The selection mechanism of `\emph{ParaDeco}$^*$'~\cite{paradeco} is based on rooted motion vector contribution. The full model does not follow the dependency chain and creates pseudo-frames using a generative predictor. Hence, the decode cost of the full model is not comparable to the cost of a pipeline where true pixels are reconstructed. We remove the generative component and retain the ranking component to create a selector we can score on the same axes as the others. The selector takes a decode budget, but in our reimplementation the contribution scores did not spread finely enough to support fine-grained budgets. We report the cheapest of them.
Both CoVA and PacketGame see the target dataset during training. \method does not.

\paragraph{Results.}
The difference between the methods lies clearly in the object detection task, and \method achieves superior results on both the VIRAT and MEVA datasets. It detects more of the moving objects from the dense detector compared to CoVA on all three datasets, $0.78$ versus $0.75$ on VIRAT, $0.60$ versus $0.54$ on ShanghaiTech, and $0.78$ versus $0.76$ on MEVA, using an equal or smaller decoding budget. 

PacketGame achieves $0.64$ on ShanghaiTech compared to our $0.60$, sending $8.8\times$ more frames to the detector ($16.8\%$ vs. $1.9\%$). PacketGame actually uses a slightly smaller decode budget compared to our method, meaning that this is a detector cost improvement rather than a decode cost improvement. ParaDeco$^*$ achieves $0.81$ on MEVA compared to our $0.78$, while decoding $3.3\times$ more frames from the GOP. On ShanghaiTech, the selector of ParaDeco$^*$ fails to outperform the keyframe-only baseline across all tasks despite still decoding $36.9\%$ of frames. The other two tasks are more dataset-dependent. On VIRAT and MEVA, BP is nearly saturated, with the keyframe-only baseline already achieving $0.99$,
whereas ShanghaiTech ranges from $0.92$ to $0.98$, with \method achieving the highest score. Counting is similar: methods are close on VIRAT and MEVA but range from $0.62$ to $0.74$ on ShanghaiTech, where \method achieves $0.74$,
tied with PacketGame. Selection therefore helps most on active content and approaches keyframe-only behavior when the scene is predictable.

\Cref{fig:filtering} shows the full recall-against-filtration rate curve. On VIRAT and MEVA our curve lies above CoVA and PacketGame at matched or lower cost. On VIRAT (the only dataset with box annotations), scoring against ground truth reproduces the same ranking. The full comparison is in the supplementary.

\subsection{System efficiency}
\label{sec:exp:latency}
The pipeline runs five stages. Entropy decoding reads the macroblock channels from the bitstream; the student produces the two maps; the selected frames and macroblocks are reconstructed; the detector runs on those frames; and motion vectors propagate the predicted bounding boxes across the remaining frames.~Decoding is only a minor component of the dense pipeline but the most costly step in the filtered one. Decoding takes up $9.6\%$ of the dense detection costs for VIRAT and $7.4\%$ for MEVA. When the detector operates only on the selected frames, the very same decoding takes up $78.5\%$ and $86.0\%$ of the costs that remain.

\begin{figure}[t]
\centering
\includegraphics[width=\columnwidth]{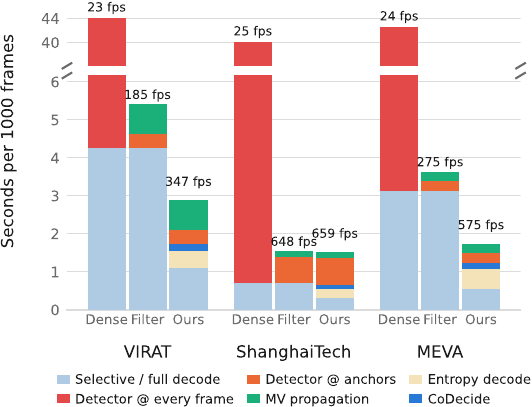}
\caption{\textbf{Where the time goes.} Wall clock times stacked per 1000 input frames, summed up for each dataset, with end-to-end throughput on top of each bar. The \emph{filter} decodes all frames and performs detection only on the selected ones, setting a limit to any selector working in the pixel space. Detection is filtered out, and the main time-consuming step is reduced by \method.}

\label{fig:throughput}

\end{figure}

\Cref{fig:throughput} illustrates two types of improvements. Filtering frames for the detector while still decoding every frame yields $8.1\times$ and $11.5\times$ on VIRAT and MEVA, respectively. \method achieves a speedup of $15.1\times$ and $24\times$, which is roughly double the gain of filtering alone.

However, the existence of such a second saving is contingent upon the relative cost of decoding to detection. The detector operates at a constant input size in our tests and thus takes $40$\,ms per frame on each of the three datasets, whereas decoding takes $4.2$, $0.7$ and $3.1$\,ms, respectively, scaling with both the image resolution and the underlying scene activity, which increases encoder complexity and subsequent decoding overhead. Consequently, ShanghaiTech (480p) starts off with $57\times$ more detection cost than decoding, so filtering already captures nearly all of the available saving and \method matches it. The reasoning runs the other way too. A faster detector shifts the balance towards decoding, which widens the gap rather than closing it. The overheads are low. Entropy decoding the video takes roughly one-tenth of full decoding, and the student runs at about $20{,}000$ frames per second. Propagation is the largest stage after decoding itself, at $27\%$ of \method's runtime on VIRAT.

\subsection{Ablations}
\label{sec:exp:ablation}
\begin{table}[t]
\centering
\small
\setlength{\tabcolsep}{3.5pt}
\begin{tabular}{@{}lcccccc@{}}
\toprule
& \multicolumn{2}{c}{VIRAT} & \multicolumn{2}{c}{ShanghaiTech} & \multicolumn{2}{c}{MEVA} \\
\cmidrule(lr){2-3}\cmidrule(lr){4-5}\cmidrule(l){6-7}
Variant & OD & Dec & OD & Dec & OD & Dec \\
\midrule
\method (ours)       & 0.78 & 29.4\% & 0.60 & 45.5\% & 0.78 & 15.5\% \\
w/o memory         & 0.66 & 4.0\% & 0.49 & 40.9\% & 0.75 & 4.9\% \\
MV only            & 0.66 & 1.2\% & 0.52 & 39.6\% & 0.73 & 1.8\% \\
NZC only           & 0.71 & 11.4\% & 0.57 & 52.0\% & 0.80 & 11.5\% \\
w/o surprise prior & 0.77 & 27.1\% & 0.60 & 46.1\% & 0.76 & \phantom{0}7.1\% \\
\bottomrule
\end{tabular}
\caption{\textbf{Component ablation.} For each variant, the decision threshold $\theta \in [0.5,1)$ is swept to match the full model's decode budget as closely as possible. Variants that cannot reach this budget are reported at their maximum coverage ($\theta = 0.5$).}
\label{tab:ablation}
\end{table}
We evaluate the contribution of each component by sequentially omitting them from the pipeline: the memory module, individual input channels, and the surprise prior. \Cref{tab:ablation} reports the results. For each ablated variant, we sweep the decision threshold to match the full model's decode budget as closely as possible. Most variants fail to reach the full budget; however, their maximum coverage is as informative as their accuracy at that operational point. The memory module proves critical for reachability. Without the ConvGRU, no threshold configuration on VIRAT or MEVA selects a comparable number of frames to the full model, because lacking a historical record prevents novelty scores from separating sufficiently. On ShanghaiTech, where the coverage does approach the baseline budget, performance degrades to $0.49$ compared to the full model's $0.60$. Similarly, neither input channel is sufficient in isolation, and the primary signal carrier depends heavily on the dataset content. While residual occupancy alone suffices on MEVA, slightly exceeding the full model at a lower budget, it underperforms on VIRAT and ShanghaiTech. Conversely, motion alone underperforms across all evaluation sets. Ultimately, combining both input channels is necessary to ensure robust performance across all three datasets.

The surprise prior is a product of channels the model already receives, so it adds no information of its own. On VIRAT and ShanghaiTech, where the variant operates at a comparable budget, it makes no difference to accuracy. This directly impacts threshold selection, which is critical for MEVA; without it, the sweep peaks at a $7.1\%$ decode rate and fails to reach the full model's budget.

Spatial selection reconstructs only the macroblocks the presence map flags. That footprint is $13\%$ of macroblocks on MEVA, $38\%$ on VIRAT, and $81\%$ on ShanghaiTech, where activity is spread across the frame. The composited frames are close to full reconstruction, at PSNR $41.96$ and SSIM $0.985$, and the saving is a further $6$ to $10\%$ of decode time per novel GOP.

\subsection{Limitations}
\label{sec:exp:limitations}
Propagation drift dominates the distance to dense detection. In VIRAT, roughly two-thirds of the missed moving boxes were properly seeded initially but drifted below IoU $0.5$ since the macroblock motion vectors are quantized, and this error builds up over a GOP. One-third were never seeded. A better propagator would do more than a better novelty head, since it is currently a major share of our cost and error. Two failure modes remain: an arriving object too small to cover enough patches for the teacher to mark it, and a novel object too small for the downstream detector.

\section{Conclusion}
\label{sec:conclusion}
In this work we show that selective video decoding can be reduced to novelty prediction in the compressed domain.~A frozen DINOv2 teacher measures novelty without any human annotation, and the codec-domain student it supervises decides,~from motion vectors, non-zero coefficient counts, and macroblock types, when and where to decode. Because supervision uses no class labels and no per-camera background models, the same trained student applies to unseen cameras and scenes. Under leave-one-dataset-out evaluation, \method matches or exceeds CoVA at equal or lower decode cost while reaching $15$ to $24\times$ dense-detection throughput. The remaining accuracy gap is dominated by propagator drift over long GOPs rather than by selection, so any better propagator can slot into the pipeline without retraining our predictor. Finally, the gain grows with the decoding-to-inference cost ratio, which faster detectors and higher resolutions both raise.

{
    \small
    \bibliographystyle{ieeenat_fullname}
    \bibliography{main}
}
\section*{Acknowledgements}
This work was supported by Milestone Research Program at AAU and Innovation Fund Denmark's grant 5304-00055B.
\clearpage
\maketitlesupplementary
\appendix
\section{Implementation Details}
\subsection*{Hyperparameters of the teacher model}
The frozen DINOv2~\cite{dinov2} teacher uses the $G{=}24$ patch grid. The L7 presence map is formed with $R_{\mathrm{THR}}{=}0.08$;
within this region, L15 Object scores above $O_{\mathrm{THR}}{=}0.20$ are
retained for the novelty stage. L23 matching uses the one-to-one cost
$c=1-\langle f_i^{23},r_j^{23}\rangle$ and marks a patch as unmatched when
$c\geq M_{\mathrm{THR}}{=}0.35$. An object is marked novel when the fraction of its constituent patches left unmatched by L23 exceeds $F_{\mathrm{THR}}{=}0.40$. Blobs with fewer than
$3$ DINO cells are not scored ($\mathrm{MIN\_BLOB}{=}3$), which suppresses
unstable single-cell responses from tiny image fragments. The reference bank is reset at each
GOP boundary and updated only after a novelty fire.

\subsection*{Student model}
Table~\ref{tab:student-spec} details the CoDecide student. The five input
channels are the four macroblock signals ($MV_x$, $MV_y$, NZC, macroblock type)
and the surprise prior
\[ s = \tanh(|MV|)\cdot\widehat{\text{NZC}}, \]
where $\widehat{\text{NZC}}$ is the per-frame max-normalized NZC channel used
as a residual proxy. All convolutions are $3{\times}3$ except the two $1{\times}1$ output
projections. GroupNorm uses 8 groups.

\begin{table}[h]
\centering\small
\begin{tabular}{lrl}
\toprule
Module & Params & Configuration \\
\midrule
Encoder            & 15{,}504  & $5{\to}32{\to}48$, GN+ReLU \\
Presence head      & 20{,}833  & $48{\to}48{\to}1$ \\
Novelty adapter    & 13{,}920  & $48{\to}32$, GN+ReLU \\
ConvGRU (mem)      & 55{,}392  & in=32, hid=32, $k{=}3$ \\
Novelty head       & 36{,}929  & $128{\to}32{\to}1$ \\
\midrule
\textbf{Total}     & \textbf{142{,}578} & \\
\bottomrule
\end{tabular}
\caption{CoDecide student parameter counts.}
\label{tab:student-spec}
\end{table}
The student is trained with the four-term objective
\begin{equation}
  \mathcal{L}=\mathcal{L}_{\mathrm{pres}}+\mathcal{L}_{\mathrm{nov}}
  +\lambda_s\,\mathcal{L}_{\mathrm{sparse}}+\lambda_g\,\mathcal{L}_{\mathrm{gop}},
\end{equation}
with $\lambda_s{=}0.2$ and $\lambda_g{=}0.3$.
The presence and novelty BCE terms have unit weights. Present-but-not-novel
cells receive a hard-negative multiplier of $5.0$, while positive BCE weights
are computed from the training-set class frequency and capped at $20$ for both
presence and novelty. The sparsity term suppresses novelty responses after an
event has been observed, and the GOP term trains the temporal keep/skip
decision.

\subsection*{Spatially selective FFmpeg reconstruction}

Our modified FFmpeg~\cite{ffmpeg} decoder separates compressed-domain parsing from pixel
reconstruction. It still entropy-decodes every macroblock to obtain motion
vectors, residual statistics, and macroblock types, but it applies a spatial mask only to the subsequent pixel-reconstruction stage. The student's $G\times G$ Presence map is upsampled with nearest-neighbor interpolation to the codec's macroblock grid. For a GOP containing a novelty event, this mask is unioned with the intra-coded macroblocks along the decoded chain, since intra blocks contain self-contained content that cannot be recovered by motion compensation.

For an ROI macroblock, FFmpeg performs the usual inverse transform,
motion-compensation, and pixel reconstruction. For a non-ROI inter-coded
macroblock, it skips those pixel operations and fills the location from the
co-located macroblock in the reference frame. The ROI mask is GOP-constant and is therefore carried through the entire I-frame-to-P-frame dependency chain; this preserves the reference pixels needed by later selected frames. The temporal decision separately sets the last reconstructed frame $f_{\mathrm{stop}}$; frames after this point are not reconstructed. Thus the method saves pixel-reconstruction work while retaining the codec's normal entropy-decode path and the dependency structure required for valid P-frame decoding.

\newcommand{\teacherrow}[1]{%
  {\setlength{\tabcolsep}{0pt}\begin{tabular}{@{}ccccc@{}}
    \includegraphics[width=0.18\textwidth]{#1/iframe.jpg} &
    \includegraphics[width=0.18\textwidth]{#1/pframe.jpg} &
    \includegraphics[width=0.18\textwidth]{#1/presence.jpg} &
    \includegraphics[width=0.18\textwidth]{#1/l15.jpg} &
    \includegraphics[width=0.18\textwidth]{#1/l23.jpg}
  \end{tabular}}\par\vspace{4pt}
}

\newcommand{\teacherheader}{%
  \noindent\makebox[0.19\textwidth][c]{\scriptsize I-frame}\hspace{0.0025\textwidth}%
  \makebox[0.18\textwidth][c]{\scriptsize P-frame}\hspace{0.0025\textwidth}%
  \makebox[0.18\textwidth][c]{\scriptsize L7 Presence map}\hspace{0.0025\textwidth}%
  \makebox[0.18\textwidth][c]{\scriptsize L15 Object map}\hspace{0.0025\textwidth}%
  \makebox[0.18\textwidth][c]{\scriptsize L23 Novelty map}\par\vspace{2pt}%
}

\begin{figure*}[t!]
  \centering
  \teacherheader
  \teacherrow{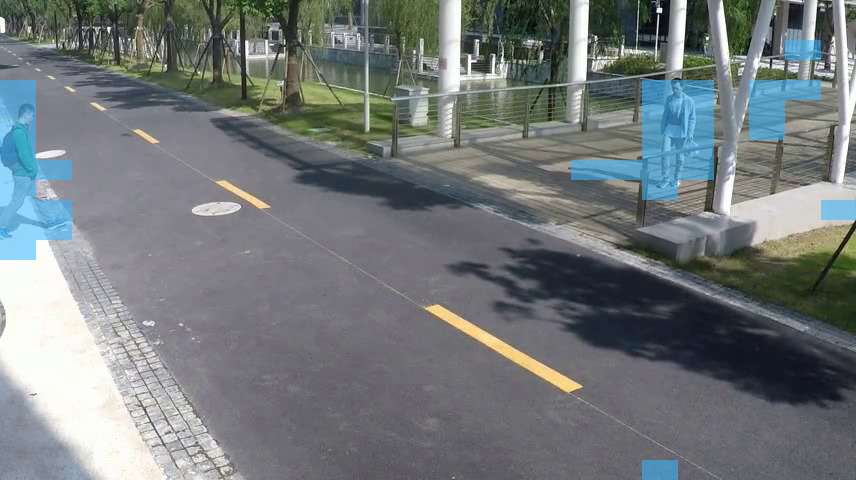}
  \teacherrow{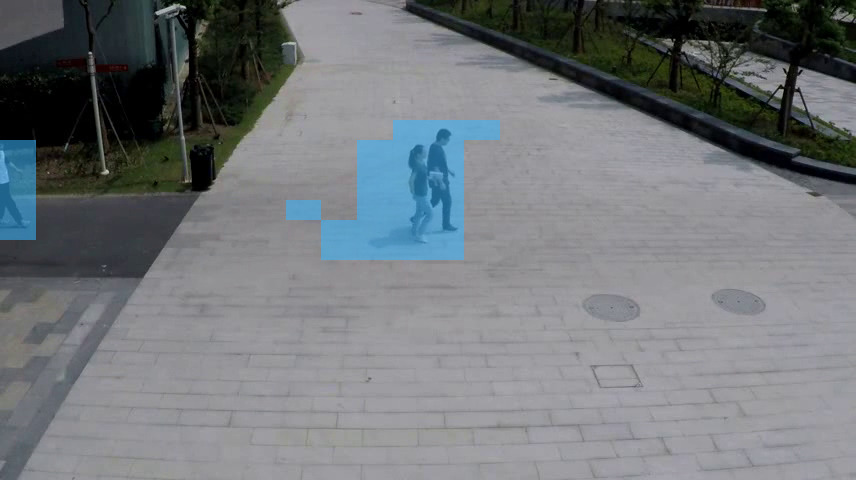}
  \teacherrow{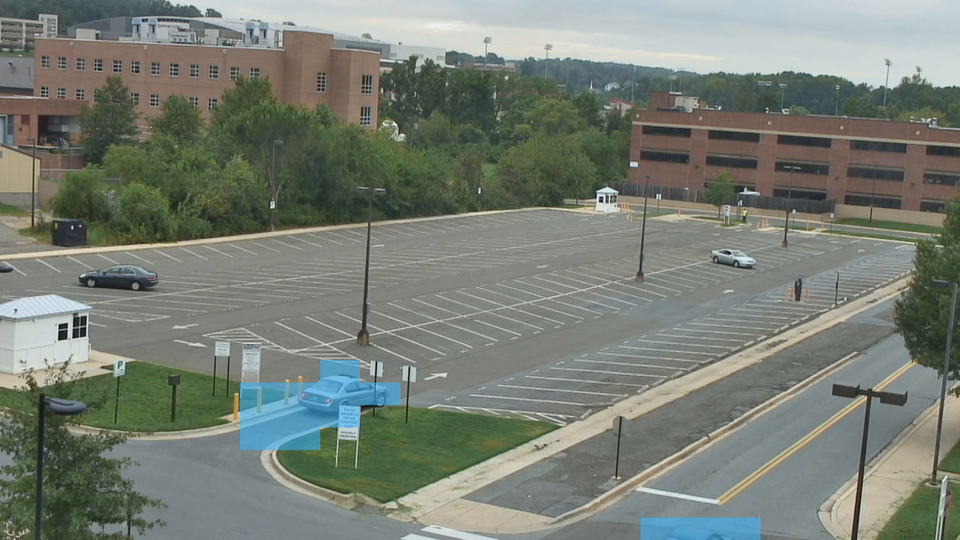}
  \teacherrow{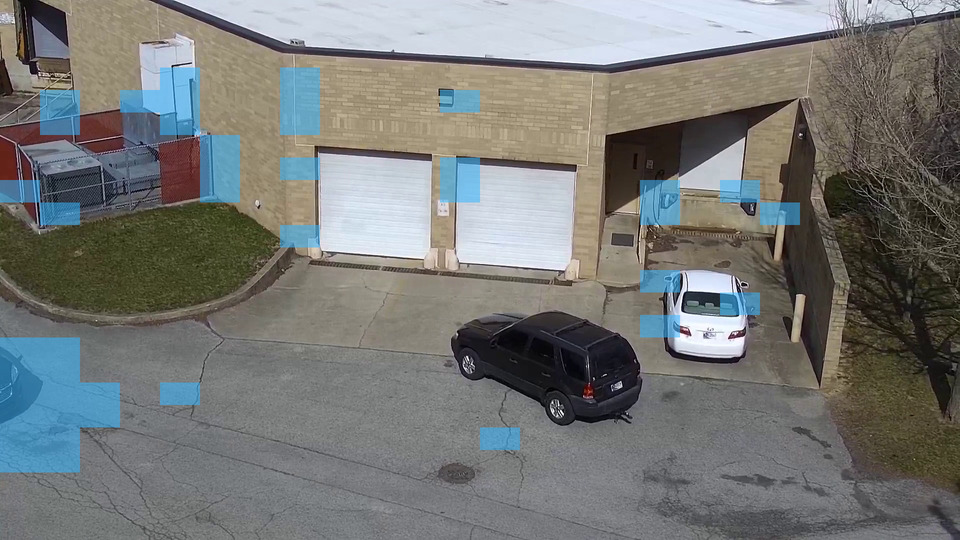}
  \teacherrow{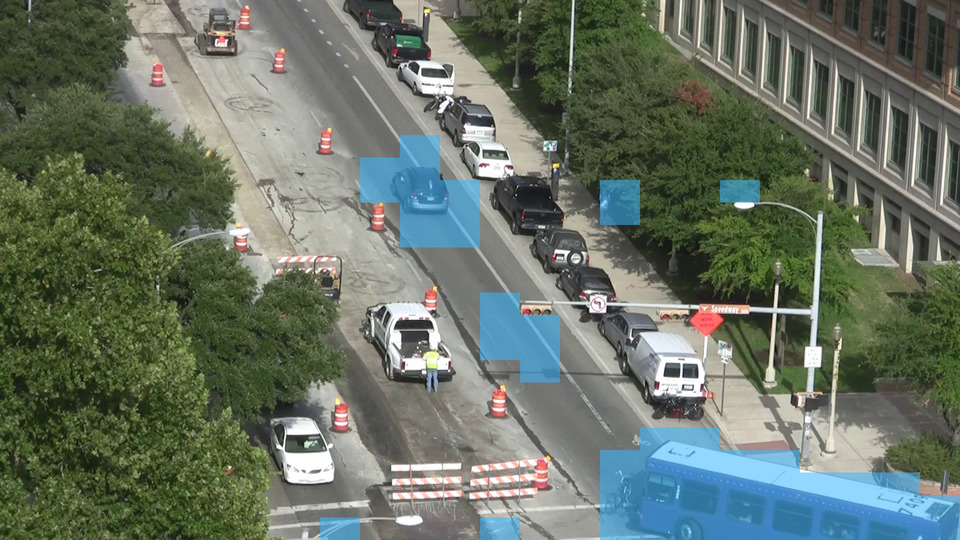}
  \caption{Qualitative teacher behavior across datasets. L7 Presence maps
  intentionally include extra motion detections beyond the target object. L15
  Object maps remove most of the background selected by Presence; Row 4 retains
  some background in its L15 map, but L23 Novelty still gives the correct result.
  Rows 1--3 show successful teacher novelty localization; Row 4 shows a newly entering car
  localized by the teacher although YOLO11x assigns it only about 0.20
  confidence because it is partially visible; Row 5 shows a bus that is
  barely visible at the I-frame boundary and becomes substantially visible
  later, with the revealed content marked as novel.}
  \label{fig:teacher-examples}
\end{figure*}
\section{Ground-truth comparison for VIRAT}
VIRAT is the only evaluated dataset with box-level human annotations. We
therefore complement the main-paper fidelity table with a direct comparison
against the VIRAT annotations in Table~\ref{tab:virat-ground-truth}. OD measures moving-object recall at IoU $=0.5$, BP is binary moving-object presence accuracy, Cnt is moving-object
counting accuracy, Sel is the fraction of frames sent to the detector, and Dec
is the fraction of the GOP reconstructed by the codec. All methods use the
same YOLO11x detector and motion-vector box propagation; only the frame
selection policy changes.
\begin{table}[t]
\centering
\small
\setlength{\tabcolsep}{3.5pt}
\begin{tabular}{l ccccc}
\toprule
& \multicolumn{5}{c}{VIRAT} \\
\cmidrule(lr){2-6}
Method & OD & BP & Cnt & Sel & Dec \\
\midrule
Keyframes only $+$ prop. & 0.54 & 1.00 & 0.85 & 0.4\% & 0.4\% \\
CoVA~\cite{cova} $+$ prop. & 0.62 & 1.00 & 0.85 & 0.7\% & 30.0\% \\
PacketGame~\cite{packetgame} $+$ prop. & 0.63 & 1.00 & 0.85 & 3.5\% & 30.0\% \\
ParaDeco$^*$~\cite{paradeco} $+$ prop. & 0.61 & 1.00 & 0.85 & 0.8\% & 51.4\% \\
\midrule
\textbf{CoDecide (ours)} & \textbf{0.64} & 1.00 & 0.85 & 1.0\% & 29.4\% \\
\bottomrule
\end{tabular}
\caption{Evaluation against ground truth on the VIRAT dataset. While Table~1 in the main text evaluates fidelity by comparing performance directly against a full YOLO per-frame detector, we present a direct comparison against the actual VIRAT human-annotated ground truth here. Each method is evaluated at the same operating point as in Table~1 of the main text.}
\label{tab:virat-ground-truth}
\end{table}

Table~\ref{tab:virat-ground-truth} provides results of VIRAT dataset against human annotated ground truth. For OD, the dense-YOLO fidelity scores and the human-GT scores are, respectively: CoDecide $0.78\rightarrow0.64$, Keyframes $0.65\rightarrow0.54$,
PacketGame $0.77\rightarrow0.63$, CoVA $0.75\rightarrow0.62$, and ParaDeco$^*$
$0.75\rightarrow0.61$. Thus CoDecide remains the best method under both
evaluations, and its advantage over keyframes is 0.13 in dense-detector
fidelity and 0.10 in human-GT OD. The absolute values diverge because the main-paper metric evaluates fidelity to the dense YOLO output, whereas the supplementary metric assesses alignment with human annotations.
The cost measurements are directly comparable: CoDecide and CoVA use 1.0\%
and 0.7\% detector-selected frames, respectively, while their chain
reconstruction costs are 29.4\% and 30.0\%. PacketGame sends 3.5\% of frames
to the detector at the same 30.0\% reconstruction cost, and ParaDeco$^*$ incurs
51.4\% reconstruction. BP and Cnt are not identical across the two tables:
Keyframes-only BP is $0.99$ against dense YOLO and $1.00$ against human GT,
and CoDecide Cnt is $0.95$ and $0.85$. This reflects differences between detector outputs and human annotations, particularly detector flicker and track/count ambiguity,
since the selected frames are identical in both evaluations.
The agreement in OD
ranking, together with the comparable Sel/Dec costs, is the intended
validation against the independent VIRAT annotations.

\section{Qualitative Teacher Examples}
This section provides qualitative examples of the DINOv2 teacher across ShanghaiTech, MEVA, and VIRAT. Figure~\ref{fig:teacher-examples} shows the same five outputs for each case: the GOP I-frame, a representative P-frame, the L7 Presence map, the L15 Object map, and the L23 Novelty map. The maps are
max-pooled over one temporal bin. Presence, Object, and Novelty use light blue, green, and red overlays, respectively.

Rows 4 and 5 show that novelty and detectability are different signals. In
Row 4, YOLO11x gives the partially occluded car only about 0.20 confidence,
yet the teacher flags it as novel, because novelty is judged by appearance
dissimilarity from the reference bank and not by detector confidence. The
frame is therefore kept, and whether the detector then recognizes the car is
a separate question. In Row 5, part of the bus is visible at the I-frame and
the rest is revealed later. L23 matches at the patch level, so the newly
uncovered pixels have no match in the bank and are marked novel, even though
the object was already partly present.

\end{document}